\documentclass{article}

\usepackage[dblblindworkshop, final]{neurips_2026}

\usepackage[utf8]{inputenc} % allow utf-8 input
\usepackage[T1]{fontenc}    % use 8-bit T1 fonts
\usepackage{hyperref}       % hyperlinks
\usepackage{url}            % simple URL typesetting
\usepackage{booktabs}       % professional-quality tables
\usepackage{amsfonts}       % blackboard math symbols
\usepackage{nicefrac}       % compact symbols for 1/2, etc.
\usepackage{microtype}      % microtypography
\usepackage{xcolor} 

\usepackage{xurl}

\usepackage{multirow} 
\usepackage{tabularx}
\usepackage{balance}
\usepackage{caption}
\usepackage{graphicx}
\usepackage[normalem]{ulem}
\usepackage[inline]{enumitem}
\usepackage{amsmath}

\usepackage{siunitx}

\newcommand{\rnd}[1]{\num{#1}}

\title{RADIUS: Ranking, Distribution, and Significance---\\A Comprehensive Alignment Suite for Survey Simulation}
\workshoptitle{Grounded User Simulation for Model Evaluation and Training: Diversity, Fidelity, and Validity}

\author{
  Weronika Łajewska \\
  Amazon \\
  Luxembourg, Luxembourg \\
  \texttt{lajewska@amazon.lu} \\
  \And
  Paul Missault \\
  Amazon \\
  Luxembourg, Luxembourg \\
  \texttt{pmissaul@amazon.lu} \\
  \AND
  George Davidson \\
  Amazon \\
  Luxembourg, Luxembourg \\
  \texttt{geodavix@amazon.lu} \\
  \And
  Saab Mansour \\
  Amazon \\
  Barcelona, Spain \\
  \texttt{saabm@amazon.es} \\
}

\begin{document}

\maketitle

\begin{abstract}
Simulating survey responses with LLMs offers a scalable alternative to costly human data collection, yet evaluation methods remain ad hoc and inconsistent. Current metrics, typically borrowed from adjacent domains, focus on accuracy or distributional similarity while overlooking ranking alignment—whether simulations preserve the relative preference orderings observed in human populations. A simulator may achieve low distributional divergence yet fail to identify the option humans most prefer, a critical gap for decision-making applications. We propose RADIUS, a two-dimensional alignment suite that evaluates survey simulation along complementary axes: 
1) \textbf{RA}nking alignment and 2) \textbf{DI}strib\textbf{U}tion alignment, accompanied by statistical \textbf{S}ignificance testing.
Experiments across multiple social surveys and persona sources demonstrate that RADIUS exposes limitations of single-metric evaluation and offers stronger discriminative power. On an independently reproduced simulator, we further show that distributional improvement often coincides with misidentifying the option humans most prefer---a decision-relevant error RADIUS surfaces but distributional metrics miss. We open-source RADIUS to support reproducible and comparable assessment of survey simulators.\footnote{\url{https://github.com/amazon-science/survey_sim_alignment_suite}}
\end{abstract}

\maketitle

\section{Introduction}
\label{sec:intro}

\begin{figure}
    \centering
    \vspace*{-0.05\baselineskip}
    \includegraphics[width=0.7\linewidth]{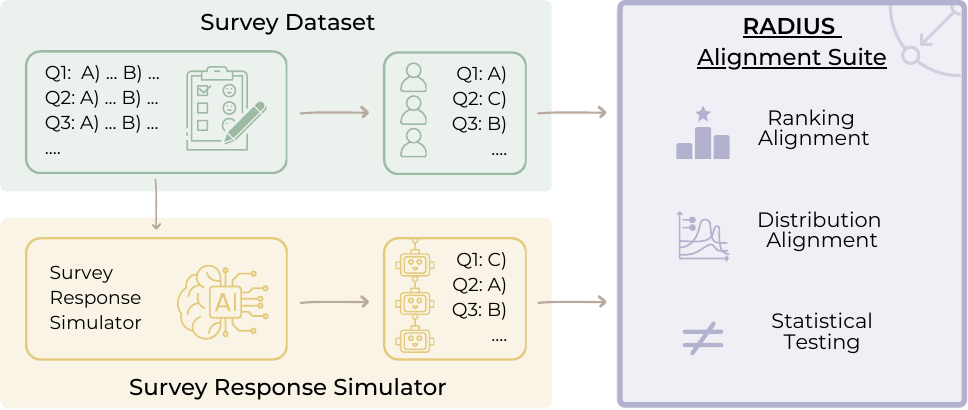}
    \captionsetup{aboveskip=8pt, belowskip=-8pt}
    \caption{User simulation framework for surveys.}
    \label{fig:sim_framework}
    \vspace{-1mm}
\end{figure}

Surveys are a cornerstone for understanding human perspectives, from large-scale social science studies to targeted e-commerce systems. However, their use is fundamentally constrained by the availability and representativeness of respondents.
This limitation has motivated a growing line of work using LLMs to simulate survey responses at scale~\citep{Gao:2023:HSSC, Park:2024:arXiv, Cao:2025:NAACL-HLT}.
Although LLMs enable agent-based simulation with controllable heterogeneity~\citep{Gao:2023:HSSC}, modeling realistic user behavior remains challenging due to the complexity of human decision-making processes. 
A general framework for LLM-based simulation includes three stages: 1) behavioral data collection to ``train'' a simulator, 2) behavioral data generation using the simulator, 3) evaluation to measure divergence from human responses (see Figure~\ref{fig:sim_framework}). We focus on the simulation alignment evaluation (stage 3) to address the lack of a standardized evaluation framework.
Survey simulation is a population-level instance of user simulation: the simulator must reproduce how a population responds rather than a single interaction. This setting offers abundant aggregated human ground truth, making it well suited to studying how faithfully simulators match real responses---provided that simulation fidelity is measured appropriately.

Most agent-based approaches simulate user behavior by assigning personas to steer model behavior~\citep{Wang:2024:FCS, Tseng:2024:EMNLP, Hu:2024:ACL}.
This shift toward persona-based simulation changes the evaluation problem: unlike earlier approaches that focused primarily on matching aggregate response distributions, persona-based LLMs generate individual-level responses that must collectively reflect population-level behavior.
Recent work has demonstrated the promise of persona-based LLMs for survey simulation~\citep{Park:2024:arXiv}, yet evaluation methods have not kept pace with this shift. Prior studies either evaluate alignment at the individual level using accuracy~\citep{Wang:2025:EMNLP_sociobench}, or at population level
using distributional metrics such as KL divergence~\citep{Namikoshi:2024:NeurIPS}, Wasserstein distance~\citep{Suh:2025:ACL}, or Cramer’s V~\citep{Argyle:2023:PA}. However, individual-level metrics require one-to-one correspondence between agents and real respondents, which is rarely available, while population-level human responses exhibit complex patterns and variability not fully captured by any single metric. 
As a result, evaluation practices vary widely across studies, with metrics chosen ad hoc, making systematic comparison difficult.
Together, these limitations underscore the need to rethink survey simulation evaluation and motivate a comprehensive alignment framework.

We propose RADIUS, a population-level alignment suite for survey simulation that defines simulation success across two complementary dimensions: 1) \emph{ranking alignment} measuring the preservation of relative option ordering and top-option alignment, and 2) \emph{distribution alignment} quantifying the probability ``mass'' shift required to match agent and human distributions and testing for statistical differences between the two.
RADIUS enables systematic quality assessment, identifies gaps in survey simulation performance in terms of different alignment dimensions, and enhances interpretability by clarifying tool capabilities for informed use.

We validate RADIUS on over 300 survey questions spanning diverse topics, including politics, family, food preferences, and religion.
We find that survey simulation alignment is multi-dimensional: ranking alignment is generally easier to achieve than distribution alignment, while distribution homogeneity is the strictest criterion, exposing even subtle mismatches. 
Our ranking metrics show higher discriminative power than commonly used alternatives, separating persona-based simulations from non-parametric baselines across most benchmarks---differences that distributional metrics alone often fail to detect.
Qualitative analysis confirms that ranking and distribution metrics expose complementary failure modes, motivating an evaluation suite that jointly considers both perspectives rather than relying on any single metric. 
Reproducing an independent simulator at scale, we further show that improving a distributional metric frequently coincides with misidentifying the human-preferred option---a verifiable, decision-relevant error that ranking evaluation catches and a distributional score alone cannot.

Overall, this paper introduces RADIUS, a multi-dimensional evaluation framework for LLM-based survey simulation. 
Our contribution is methodological: we synthesize complementary metrics---including Top Rank Match (TRM), a new tie-aware top-choice agreement metric---into a unified suite for measuring population-level simulation fidelity, demonstrate its discriminative power over single-score evaluation across domains and persona sources, and provide an open-source implementation for reproducible assessment. 

\section{Related Work}

User simulation uses intelligent agents to mimic human interactions with systems~\citep{Balog:2024:FnTIR_user_sim}, supporting applications such as automatic evaluation~\citep{Jung:2009:CSL, Abbasiantaeb:2024:WSDM}, behavioral analysis~\citep{Hazrati:2024:UMUAI, Park:2024:arXiv}, experiment piloting~\citep{Filippas:2024:EC}, social scenario simulation~\citep{Gao:2024:WWW, Park:2022:UIST}, and synthetic data generation~\citep{Balog:2025:SIGIR}. Effective simulation requires validity, cognitive plausibility, and behavioral variation~\citep{Balog:2024:FnTIR_user_sim, Bighold:2021:MDPI}. Recent advances in LLMs have significantly improved the realism of user simulators~\citep{Wang:2025:TOIS, Argyle:2023:PA, Aher:2023:ICML}, enabling agents to assume specific roles through prompting~\citep{Wang:2024:FCS, Tseng:2024:EMNLP, Hu:2024:ACL, Park:2024:arXiv}. These roles are typically defined by user profiles that include demographic attributes, personality traits, and social context tailored to the target application~\citep{Chen:2024:TMLR, Wang:2024:FCS}.

User simulation is particularly valuable in social science research, where collecting human responses is costly. \citet{Park:2024:arXiv} showed that generative agents can reproduce survey responses with 85\% accuracy, comparable to participants’ own two-week consistency.
Existing approaches include persona-based response generation~\citep{Park:2024:arXiv} and LLMs fine-tuned for multiple-choice questions~\citep{Cao:2025:NAACL-HLT}. Simulators may operate at the individual level to replicate specific user behaviors~\citep{Park:2024:arXiv, Wang:2025:EMNLP_sociobench}, or at the population level to generate response distributions for targeted sub-populations~\citep{Argyle:2023:PA, Cao:2025:NAACL-HLT, Suh:2025:ACL}. While surveys span diverse formats, including Likert-scale, multi-select, ranking, and open-ended questions, this work focuses on multiple-choice single-select questions to enable more precise modeling and evaluation, leaving open-ended survey simulation to future work~\citep{Ma:2025:ACL, Ma:2025:arXiv}.

Despite promising results, replicating the complexity of real human opinions with LLM-based agents remains challenging~\citep{Kaiser:2025:UMAP}, and simulated responses often show less variation than real survey data~\citep{Bisbee:2024:PA}. A central unresolved question is how to evaluate the alignment between simulated and human responses. Survey simulation alignment is influenced by multiple factors, including the question domain~\citep{Meister:2025:NAACL-HLT}, persona design~\citep{Lutz:2025:EMNLP}, steering methods~\citep{Meister:2025:NAACL-HLT, Bui:2025:ACL}, and response representation (e.g., single choice, distributions, log-probabilities)~\citep{Meister:2025:NAACL-HLT}. 
Given the many factors affecting simulation alignment, developing consistent evaluation methods is challenging. Existing approaches fall into two categories. Individual-level alignment relies on metrics such as accuracy~\citep{Park:2024:arXiv, Wang:2025:EMNLP_sociobench} and RMSE~\citep{Namikoshi:2024:NeurIPS}. 
Population-level alignment uses distributional measures, including Total Variation Distance (TVD)~\citep{Hu:2025:ACL}, SimBench score (SBS)~\citep{Hu:2025:ACL} (TVD normalized against a uniform baseline), Kullback–Leibler divergence (KLD)~\citep{Namikoshi:2024:NeurIPS}, Jensen–Shannon divergence (JSD)~\citep{Cao:2025:NAACL-HLT}, Wasserstein distance (WD)~\citep{Cao:2025:NAACL-HLT, Suh:2025:ACL}, Cramer’s V correlation (CV)~\citep{Argyle:2023:PA}, ANOVA~\citep{Park:2024:arXiv, Kaiser:2025:UMAP}, and model log-probabilities~\citep{Meister:2025:NAACL-HLT}. 
The prior work closest to our top-choice metric reports accuracy against the single human majority option~\citep{Cao:2025:NAACL-HLT}, but without accounting for statistically tied top options; separately, distribution-level tests assess whether human and simulated distributions differ as a whole~\citep{Sun:2024:arXiv}. Our TRM is tie-aware, crediting any human top option statistically indistinguishable from the most-voted one via bootstrap grouping.
Each of these metrics captures only partial aspects of alignment, and the lack of standardization hinders comparison across studies. We address this gap by proposing a dedicated alignment suite for evaluating survey simulations.
\section{Survey Simulation}

Since our goal is establishing a reusable evaluation framework rather than improving simulation itself, we employ a standard simulation environment with persona-based prompting, leaving more sophisticated approaches for future work~\citep{Tseng:2024:EMNLP, Chen:2024:TMLR, Hu:2024:ACL}. 
The core component is a population of agents generated from behavioral data. To assess how persona source affects alignment, we construct personas from three distinct domains: (1)~social science survey responses (Soc)~\citep{Kolluri:2025:EMNLP}, providing  attitudinal and demographic grounding; (2)~Rotten Tomatoes movie critic reviews (RT),~\footnote{\url{https://www.kaggle.com/datasets/stefanoleone992/rotten-tomatoes-movies-and-critic-reviews-dataset}} capturing cultural preferences and judgment patterns; and (3)~e-commerce purchase histories (E-c)~\citep{Berke:2023:HarvardDataverse}, reflecting consumer behavior.
Each agent's persona encompasses demographics, interests, and behavioral patterns~\citep{Lutz:2025:EMNLP}, with the specific composition determined by the source domain.
Personas derived from behavioral data provide context for an LLM to role-play an individual with specific traits~\citep{Chen:2024:TMLR}, selecting the response that best aligns with portrayed character.\footnote{The simulation setup serves as a testbed for the alignment suite rather than a contribution in itself; our focus is on evaluation methodology, not simulator optimization.} We use Claude Sonnet 4.5~\citep{Anthropic:2025:Sonnet4.5} as the simulator across all persona sources and benchmarks.\footnote{\texttt{anthropic.claude-sonnet-4-5-20250929} via Amazon Bedrock, decoding temperature $0$ for reproducible outputs; each of the 1{,}000 persona-based agents per source answers independently, invalid answers are discarded, and per-question response distributions are aggregated across the persona population.} Since the alignment suite is model-agnostic, it applies unchanged to any simulator that produces response distributions.
\section{RADIUS Alignment Suite}

Our evaluation approach goes beyond identifying the ``winner'' option to assess whether simulated surveys capture the patterns and variability of human responses. The RADIUS alignment suite operates at multiple levels: question-level metrics compare response rankings and distributions, survey-level aggregations synthesize overall alignment, and statistical tests quantify the significance of differences. The ultimate goal is to ensure that simulated data reflects both the general human patterns and natural individual variability. Alignment is defined via two complementary criteria:
\begin{itemize}[leftmargin=1em]
    \item \emph{Ranking alignment} measuring how well the relative ordering of options is preserved and verifying the top-choice alignment.
    \item \emph{Distribution alignment} quantifying the probability mass needed to transform agent distributions into human distributions and testing for the statistically significant differences between them.
\end{itemize}

\subsection{Ranking Alignment}

We evaluate ranking alignment at two levels: 1) top-choice agreement between humans and agents, and 2) agreement in the relative ordering of all answer options.

\paragraph{Top Rank Match (TRM)}

TRM quantifies whether agents correctly identify the top-ranked human preferences, accounting for statistical significance in human votes. 
It is binary, with 1 indicating perfect alignment with human top choice within statistical bounds. The metric recognizes that when humans cannot distinguish between their first and second choices, agents should not be penalized for selecting the second-ranked option. Conversely, when humans show a significant preference for their top choice over their second choice, agents are penalized for predicting the second-ranked option. 
To take confidence of human top choice into account, we create option groups using confidence intervals identified with bootstrap significance testing~\citep{Efron:1994:Chapman_bootstrap}, which is a model-free approach that extends to all types of questions and setups.
For each question's human vote distribution, we perform bootstrap resampling ($n=1000$) to determine confidence intervals and identify statistically tied options. 
Next, we identify the top option group by clustering options whose confidence intervals overlap with the highest-voted option.
Finally, we evaluate binary top rank alignment where the simulator prediction is considered correct if its top choice falls within the human top choice option group. 
This metric is intuitive and easily interpretable, handles ties in human preferences via bootstrap confidence intervals, supports per-question analysis with option groups, and is robust to small sample sizes through bootstrapping.
TRM generalizes the majority-choice accuracy used in prior work~\citep{Cao:2025:NAACL-HLT}: rather than binary top-1 agreement with the single most-voted human option, it credits any statistically indistinguishable top option via bootstrap grouping, avoiding unfair penalties when human first and second choices are tied.

\paragraph{Rank Correlation (RC)}

RC quantifies the alignment between the option ordering induced by agent responses and that induced by human responses.
We implement this metric using the Spearman correlation coefficient, which measures the strength and direction of monotonic relationships between two ranked variables. 
The coefficient is calculated as: $RC = 1 - \frac{6\sum d^2}{n(n^2-1)}$, where $d$ represents the absolute difference between ranks for each pair of data points and $n$ is the total number of response options.
The Spearman correlation coefficient analyzes the ranks of data points rather than raw values, making it particularly suitable for non-normally distributed data, ordinal data, or datasets containing outliers. To facilitate the interpretation of the results, we normalize the coefficient ($RC_{norm} = \frac{RC+1}{2}$), resulting in a scale where 1 indicates perfect positive correlation and 0 indicates no correlation between human and agent rankings.
Compared to TRM, RC provides a softer measure of rank distance, with larger penalties for greater rank discrepancies from human responses.

\subsection{Distribution Alignment}

While ranking metrics capture the correct ordering of options, they are insensitive to prevalence: the distributions [0.51, 0.49] and [0.99, 0.01] have perfect ranking scores despite representing very different response frequencies. Distribution alignment metrics fill this gap.

\paragraph{Total Variation Distance (TVD)}

TVD quantifies the minimum percentage of probability mass that needs to be redistributed to make two response distributions identical, calculated as: $\text{TVD} = \frac{1}{2}\sum_{i=1}^{n}|H_i - A_i|$, where $H$ represents the human response distribution, $A$ represents the simulated response distribution, and $n$ is the number of response options~\citep{Hu:2025:ACL}.
In principle, TVD represents the maximum possible difference between the probability of any event occurring under two different distributions. It ranges from 0 to 1, with lower values indicating better simulation fidelity. Unlike alternatives such as KL divergence, TVD remains well-defined when models assign zero probability to human responses, while also offering symmetry and boundedness for improved interpretability across datasets~\citep{Hu:2025:ACL}. 
TVD is highly sensitive to truly categorical questions and large shifts in popular options, but less sensitive to ordered responses or small changes spread across many options.

\paragraph{Distribution Homogeneity (DH)}

The chi-square test of homogeneity assesses whether categorical response distributions are the same across different populations (agents vs humans)~\citep{Sun:2024:arXiv}.
The null hypothesis assumes the same distributions across populations, while the alternative hypothesis states they differ. 
Chi-squared test p-values below 0.05 indicate a statistically significant difference between the distributions. We report DH as a binary value, with 0 indicating significant distributional mismatch for a given question. 
DH is complementary to TVD, as it tests whether simulated and human response distributions are statistically distinguishable rather than measuring the magnitude of their difference.

\subsection{Survey-Level Aggregation and Significance Testing}

Survey-level alignment is computed by averaging question-level scores for each metric. To assess differences between simulator runs, we apply paired t-tests to question-level scores,
which naturally pair Simulator A and Simulator B responses for each question, testing whether the mean difference is significantly different from zero.
\begin{table}[t]
      \footnotesize
      \centering
      \setlength{\tabcolsep}{4pt}
      \caption{Survey-level simulation alignment across persona sources (Soc, RT, E-c), test survey benchmarks (GSS, W50-92), and non-parametric baseline (Nor). \underline{Best} per benchmark; * = significantly different from best ($p < 0.05$, paired  t-test).}
      \begin{tabular}{lllllllllllll}
      \toprule
      & \multicolumn{4}{c}{GSS} & \multicolumn{4}{c}{W50} & \multicolumn{4}{c}{W92} \\
      \cmidrule(lr){2-5} \cmidrule(lr){6-9} \cmidrule(lr){10-13}
      Metric & Soc & RT & E-c & Nor & Soc & RT & E-c & Nor & Soc & RT & E-c & Nor \\
      \midrule
      \multicolumn{13}{l}{\textit{RADIUS (proposed alignment suite)}} \\
      TRM $\uparrow$ & \rnd{0.700} & \rnd{0.667} & \underline{\rnd{0.767}} & \rnd{0.450}* & \rnd{0.352}* & \rnd{0.273}* & \underline{\rnd{0.484}} & \rnd{0.391} & \underline{\rnd{0.649}} & \rnd{0.494}* & \rnd{0.584} & \rnd{0.364}* \\
      RC $\uparrow$ & \rnd{0.820} & \rnd{0.764} & \underline{\rnd{0.833}} & \rnd{0.613}* & \rnd{0.534}* & \rnd{0.487}* & \rnd{0.623}* & \underline{\rnd{0.710}} & \underline{\rnd{0.871}} & \rnd{0.675}* &
  \rnd{0.756}* & \rnd{0.675}* \\
      TVD $\downarrow$ & \rnd{0.305} & \rnd{0.379}* & \rnd{0.344}* & \underline{\rnd{0.252}} & \rnd{0.574}* & \rnd{0.657}* & \rnd{0.494}* & \underline{\rnd{0.380}} & \underline{\rnd{0.278}} & \rnd{0.455}* &
  \rnd{0.451}* & \rnd{0.306} \\
      DH $\uparrow$ & \underline{\rnd{0.033}} & \rnd{0.000} & \rnd{0.017} & \rnd{0.033} & \rnd{0.000} & \rnd{0.000} & \underline{\rnd{0.016}} & \rnd{0.000} & \underline{\rnd{0.000}} & \rnd{0.000} & \rnd{0.000} &
   \rnd{0.000} \\
      \midrule
      \multicolumn{13}{l}{\textit{Prior work (for comparison)}} \\
      CV $\downarrow$ & \rnd{0.321} & \rnd{0.408}* & \rnd{0.374}* & \underline{\rnd{0.270}} & \rnd{0.594}* & \rnd{0.667}* & \rnd{0.524}* & \underline{\rnd{0.424}} & \underline{\rnd{0.206}} & \rnd{0.369}* &
  \rnd{0.337}* & \rnd{0.348}* \\
      JSD $\downarrow$ & \rnd{0.089}* & \rnd{0.132}* & \rnd{0.118}* & \underline{\rnd{0.054}} & \rnd{0.306}* & \rnd{0.370}* & \rnd{0.243}* & \underline{\rnd{0.126}} & \underline{\rnd{0.073}} & \rnd{0.190}* &
  \rnd{0.183}* & \rnd{0.094}* \\
      WD $\downarrow$ & \underline{\rnd{0.378}} & \rnd{0.550}* & \rnd{0.490}* & \rnd{0.379} & \rnd{1.104}* & \rnd{1.326}* & \rnd{0.910}* & \underline{\rnd{0.688}} & \underline{\rnd{0.350}} & \rnd{0.698}* &
  \rnd{0.635}* & \rnd{0.568}* \\
      % \midrule
      SBS $\uparrow$ & -64.7 & -119.2 & -98.5 & -0.1 & -55.2 & -76.2 & -30.4 & 7.4 & 10.4 & -49.2 & -45.7 & 7.8 \\
      \bottomrule
      \end{tabular}
      \label{tab:results}
      \vspace{-1mm}
  \end{table}
\begin{table*}[t]
  \setlength{\tabcolsep}{2pt}
  \footnotesize
  \centering
  \caption{Example questions illustrating complementary failure modes, with human and agent response counts and computed metrics. \emph{Src} denotes the persona source (Soc: social science, RT: Rotten   Tomatoes, E-c: e-commerce) and survey benchmark (GSS, W34/W50/W92 OpinionQA waves).}
  \begin{tabular}{p{4.8cm} l l ccccc cc}
      \toprule
      & & & \multicolumn{4}{c}{RADIUS} & \multicolumn{3}{c}{Prior work} \\
      \cmidrule(lr){4-7} \cmidrule(lr){8-10}
      Question & Src & Human/Agent Counts & TRM & RC & TVD & DH & WD & CV & JSD \\
      \midrule
      \multirow{2}{4.8cm}{1. Please tell us whether you are satisfied or dissatisfied with your social life.} & \multirow{2}{*}{Soc/W50}
        & {\scriptsize [3011, 4590, 1770, 447, 16]} & \multirow{2}{*}{\rnd{1.00}} & \multirow{2}{*}{\rnd{0.65}} & \multirow{2}{*}{\rnd{0.51}} & \multirow{2}{*}{\rnd{0.00}} & \multirow{2}{*}{\rnd{0.52}} & \multirow{2}{*}{\rnd{0.34}} & \multirow{2}{*}{\rnd{0.21}} \\
        & & {\scriptsize [1, 919, 22, 0, 57]} & & & & & & & \\
      \addlinespace
      \multirow{2}{4.8cm}{2. How often do you choose foods to eat because they are healthy and nutritious} & \multirow{2}{*}{RT/W34}
        & {\scriptsize [310, 1024, 784, 368, 46, 5]} & \multirow{2}{*}{\rnd{0.00}} & \multirow{2}{*}{\rnd{0.96}} & \multirow{2}{*}{\rnd{0.19}} & \multirow{2}{*}{\rnd{0.00}} & \multirow{2}{*}{\rnd{0.33}} & \multirow{2}{*}{\rnd{0.23}} & \multirow{2}{*}{\rnd{0.05}} \\
        & & {\scriptsize [5, 354, 453, 188, 0, 0]} & & & & & & & \\
      \addlinespace
      \multirow{3}{4.8cm}{3. Are we spending too much, too little, or about the right amount on solving the problems of big cities?} & \multirow{3}{*}{RT/GSS}
        & {\scriptsize [260, 410, 921]} & \multirow{3}{*}{\rnd{1.00}} & \multirow{3}{*}{\rnd{1.00}} & \multirow{3}{*}{\rnd{0.41}} & \multirow{3}{*}{\rnd{0.00}} & \multirow{3}{*}{\rnd{0.56}} & \multirow{3}{*}{\rnd{0.45}} & \multirow{3}{*}{\rnd{0.14}} \\
        & & {\scriptsize [5, 11, 984]} & & & & & & & \\
        & & & & & & & & & \\
      \addlinespace
      \multirow{2}{4.8cm}{4. What about sexual relations between two adults of the same sex--do you think it is...} & \multirow{2}{*}{E-c/GSS}
        & {\scriptsize [741, 84, 137, 1163]} & \multirow{2}{*}{\rnd{1.00}} & \multirow{2}{*}{\rnd{0.97}} & \multirow{2}{*}{\rnd{0.45}} & \multirow{2}{*}{\rnd{0.00}} & \multirow{2}{*}{\rnd{1.18}} & \multirow{2}{*}{\rnd{0.46}} & \multirow{2}{*}{\rnd{0.18}} \\
        & & {\scriptsize [2, 0, 0, 998]} & & & & & & & \\
      \addlinespace
      \multirow{2}{4.8cm}{5. Do you restrict or limit eating sugar on a regular basis?} & \multirow{2}{*}{E-c/W34}
        & {\scriptsize [1455, 1082]} & \multirow{2}{*}{\rnd{0.00}} & \multirow{2}{*}{\rnd{0.00}} & \multirow{2}{*}{\rnd{0.23}} & \multirow{2}{*}{\rnd{0.00}} & \multirow{2}{*}{\rnd{0.23}} & \multirow{2}{*}{\rnd{0.21}} & \multirow{2}{*}{\rnd{0.03}} \\
        & & {\scriptsize [342, 658]} & & & & & & & \\
      \addlinespace
      \multirow{2}{4.8cm}{6. Which statement comes closer to your view?} & \multirow{2}{*}{Soc/W50}
        & {\scriptsize [2603, 2256, 48]} & \multirow{2}{*}{\rnd{0.00}} & \multirow{2}{*}{\rnd{0.75}} & \multirow{2}{*}{\rnd{0.08}} & \multirow{2}{*}{\rnd{0.00}} & \multirow{2}{*}{\rnd{0.08}} & \multirow{2}{*}{\rnd{0.06}} & \multirow{2}{*}{\rnd{0.00}} \\
        & & {\scriptsize [449, 537, 13]} & & & & & & & \\
      \bottomrule
  \end{tabular}
  \label{tab:qualitative_analysis_examples}
\end{table*}

\section{Results}
\label{sec:results}

To validate the sensitivity of our metrics, we evaluate simulated survey responses using RADIUS alongside metrics from prior work (Table~\ref{tab:results}) across surveys spanning diverse domains and question complexity.
Our test set includes the General Social Survey (GSS; political views/religion/science/crime) and two waves of OpinionQA~\citep{Shibani:2023:ICML_opinionqa}: W50 (relationships/family) and W92 (political views).\footnote{Additional results on W34 (biomedical/food) are consistent with reported findings.}

\subsection{Relative Strictness of Alignment Criteria}

We observe a clear hierarchy in alignment difficulty: ranking alignment is generally easier to achieve than distribution alignment, while DH is the most stringent criterion, requiring near-perfect agreement with no statistical differences. In contrast, TVD captures the magnitude of disagreement between distributions, providing a graded assessment even when differences are statistically significant.
Ranking-based metrics are less strict because they depend only on the relative ordering of response options rather than their exact probability mass. As a result, substantial distributional mismatches can coexist with preserved rankings between human and agent responses. Among ranking metrics, RC is easier to satisfy than TRM: high correlation can be achieved even when the most preferred option differs, whereas top-rank agreement requires exact matching of the highest-ranked choice. Together, these results highlight the complementary roles and differing strictness of alignment metrics, motivating the need for a multi-dimensional evaluation framework.

\subsection{Discriminative Power of Metrics}

Alignment varies substantially by topic, question complexity, and persona source.
On W92, all RADIUS metrics identify Soc as the best configuration (or tied on DH). In contrast, on GSS and W50 no single configuration is best across all metrics:
persona sources achieve the best top-rank alignment (TRM) while the normal baseline achieves the best distributional alignment.
A single-score metric would collapse this tradeoff into a misleading verdict: the SimBench score~\citep{Hu:2025:ACL}, which measures distributional improvement over a uniform baseline, is negative for 8 of 9 persona configurations, meaning distributions are further from human responses than uniform guessing, despite strong ranking alignment (TRM up to 0.77, RC up to 0.87). RADIUS exposes this tradeoff, enabling informed choices based on application requirements---a distinction that single-score evaluations would obscure.

\subsection{Alignment Metrics Robustness}

To assess metric robustness, we compare persona-based simulations against a normal distribution baseline (Nor), which we found consistently outperforms random (Dirich\-let-Multinomial) and uniform alternatives (full results for all three baselines in Appendix~\ref{sec:appendix_baselines}).
Alignment scores from this baseline are compared to persona-based LLM simulations, testing whether the non-parametric responses are statistically distinguishable from those conditioned in personas. 
Overall, our ranking metrics (TRM, RC) assign significantly worse scores to baselines than to persona-based simulations on GSS and W92, while commonly used distribution metrics (JSD, CV) often show no significant advantage for persona sources or even favor the baselines.
For surveys where LLM simulations are misaligned (W50), the normal baseline sometimes scores better, highlighting cases where LLMs do not outperform trivial baselines.
Distribution alignment metrics often favor non-parametric baselines: on GSS and W50, normal distributions achieve lower TVD and JSD than all persona sources, indicating better distributional shape despite the baseline being content-agnostic.
This pattern reflects the \emph{alignment-simulation tradeoff} identified by~\citet{Hu:2025:ACL}: instruction-tuned models used for persona role-playing produce confident, decisive outputs, causing mode collapse---concentrating responses on one or two options rather than reproducing the full diversity of human opinion.
Together, these results confirm that RADIUS provides more robust discrimination between meaningful simulation and statistical noise than commonly used distributional measures alone.

\subsection{Metric Complementarity}
\label{subsec:complementarity}

We qualitatively examine question scores to assess whether ranking- and distribution-based metrics capture complementary failure modes in survey simulation (see Table~\ref{tab:qualitative_analysis_examples}; examples span all evaluated benchmarks including W34).
Across questions, we observe that strong performance under one metric can mask substantial misalignment under another, underscoring the need for multi-dimensional evaluation.
In some cases, agreement on the most frequent response masks severe distributional mismatch. For Q1, the simulator achieves perfect TRM (1.00), but moderate RC (0.65) and poor distribution alignment (TVD = 0.51). While both humans and agents most often select the 2nd option, the agent concentrates probability on a narrow subset of options versus the much broader human distribution---a manifestation of mode collapse.
Conversely, low top-rank agreement does not necessarily imply poor overall alignment. For Q2, TRM is 0.00, while RC is high (0.96) and TVD is low (0.19), reflecting only a minor swap among the top options rather than significant disagreement in preferences.
On the other hand, in Q3, both TRM and RC are perfect (1.00), yet distribution alignment remains poor (TVD = 0.41), indicating that identical orderings can coexist with distribution mismatch.
Standard distributional metrics can also give misleading signals in both directions. For Q4, ranking alignment is near-perfect (TRM = 1.00, RC = 0.97), yet WD is high (1.18) as the agent correctly identifies the top option but concentrates nearly all probability mass on it.
In the opposite direction, standard distributional metrics can miss ranking errors. For Q5, the agent completely flips the human preference ordering (TRM = 0), yet JSD remains very low (0.03), failing to flag a clear ranking disagreement.
Similarly, in Q6, CV is low (0.06) despite the agent selecting the wrong top option (TRM = 0). 
Notably, DH is 0.00 for all examples, including Q6 where TVD and JSD indicate near-perfect distributional match, demonstrating that DH is the strictest criterion, detecting statistically significant differences even when the magnitude of mismatch is negligible.
Altogether, these examples show that ranking and distribution metrics surface distinct and complementary aspects of survey alignment, and that relying on any single metric can lead to misleading conclusions.

\begin{figure}[t]
    \centering
    \includegraphics[width=0.5\linewidth]{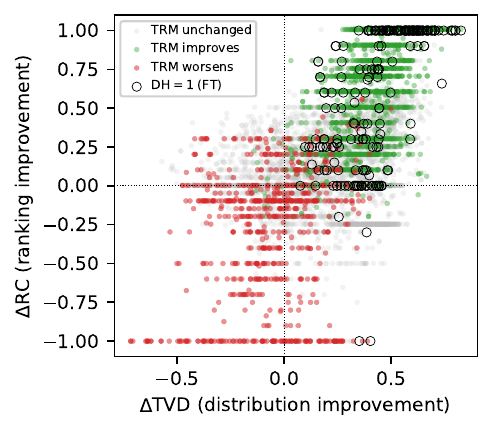}
    \caption{Per-question change after fine-tuning on the \citet{Cao:2025:NAACL-HLT} simulator, evaluated with RADIUS. Each point is one question: $x$ = distribution improvement ($\Delta$TVD), $y$ = ranking improvement ($\Delta$RC); colour marks the change in top-rank agreement (TRM), and thin circles mark questions reaching distributional homogeneity (DH$=$1) after fine-tuning.}
    \label{fig:cao_scatter}
\end{figure}

\subsection{RADIUS Beyond Our Experimental Setup}
\label{subsec:external_complementarity}

The analyses above establish metric complementarity within our own experimental setup (Section~\ref{subsec:complementarity}). To test whether it holds at scale and beyond our own design choices, we reuse the large-scale World Values Survey data and reproduce a fine-tuned simulator of \citet{Cao:2025:NAACL-HLT}---an independently trained model, dataset, and method not designed by us---as a second testbed, and evaluate it with RADIUS.\footnote{Reproduction details, including our fidelity check against their published scores, are in Appendix~\ref{sec:appendix_reproducibility}.} Figure~\ref{fig:cao_scatter} plots, per question, the change in distribution alignment ($\Delta$TVD) against the change in ranking alignment ($\Delta$RC) after fine-tuning, with the change in TRM shown by colour and questions reaching distributional homogeneity (DH$=$1) marked. The distribution and ranking axes are correlated but far from equivalent: fine-tuning improves TVD on 82.5\% of questions, yet 67.7\% of those show no improvement in TRM and 4.9\% show TRM \emph{worsening}, while DH is attained almost only where both axes improve. 

Crucially, this divergence corresponds to real, verifiable errors rather than mere metric disagreement. Among the questions whose TVD improves after fine-tuning, 45.9\% still assign the top rank to an option other than the human population's most-preferred one (TRM$=$0, verified against the ground-truth response distribution), and 21.4\% of all questions do so despite appearing well-aligned distributionally (TVD $<$ 0.3). A practitioner relying on distributional improvement alone would accept these simulators as improved, yet they misidentify the option humans most prefer---precisely the decision-relevant error RADIUS's ranking axis surfaces and a single distributional score cannot. This confirms, on data and a simulator not designed by us, that RADIUS's dimensions capture distinct and decision-relevant information.
\section{Conclusions}

We introduced RADIUS, a population-level evaluation framework for survey simulation that jointly captures ranking alignment, distribution alignment, and statistical significance through complementary, interpretable metrics supporting discriminative comparison between approaches. Our analyses show that alignment is inherently multi-faceted: ranking and distribution metrics capture related but distinct information, and relying on either alone yields a misleading verdict---persona-based simulations can preserve human preference structure while exhibiting mode collapse in their distributions, or match a distribution's shape while misidentifying the most-preferred option. We observe this complementarity both across our own persona sources and benchmarks and, independently, when reproducing a fine-tuned simulator not designed by us, where improving distributional alignment frequently leaves the human-preferred option misidentified---an error verifiable against real responses that ranking evaluation catches. By framing survey alignment as a multi-dimensional construct, RADIUS provides a foundation for systematically assessing and comparing the population-level fidelity of survey simulation methods. Future work could investigate the relationship between individual- and group-level alignment and develop simulation approaches that mitigate mode collapse while preserving preference structure.

\section*{Limitations}
  
Our evaluation focuses on multiple-choice single-select questions---the format with the broadest public ground-truth coverage (GSS, OpinionQA). This reflects data availability rather than a restriction of the metrics: TRM's bootstrap option grouping is model-free and applies to any categorical response options set (including multi-select), and RC uses Spearman rank correlation, which is defined on ordinal data and already rewards ``close'' answers on Likert scales, penalizing adjacent transpositions less than distant ones. Distribution metrics that treat options as unordered (TVD, DH) can be extended to ordinal scales via distance-weighted variants such as the WD we already report, while open-ended responses would require a prior discretization or embedding step. 
TRM's tie rule based on  overlapping confidence intervals is conservative (overlap does not imply a non-significant difference), and tie groups depend on human sample size; testing the top-option difference directly is a natural refinement. We also do not study prompt variants or repeated runs, which may affect absolute scores.
Finally, our primary experiments use a single prompt-based simulator with personas derived from specific behavioral datasets; although we additionally evaluate an independently fine-tuned simulator (Section~\ref{subsec:external_complementarity}), alignment scores in general depend on both persona design and the underlying model, and absolute values may shift under alternative simulation setups. As our contribution is the evaluation framework rather than the simulator, we expect the relative, multi-dimensional comparisons RADIUS enables to remain informative across such choices.

% \balance
\bibliographystyle{plainnat}
\bibliography{references}

\appendix
\section{Results on OpinionQA W34}
\label{sec:appendix_w34}

For space, the main text reports GSS, W50, and W92 (Table~\ref{tab:results}). Table~\ref{tab:appendix_w34} reports the additional OpinionQA W34 wave (biomedical and food-habit topics), across all three persona sources (Soc, RT, E-c) and all three non-parametric baselines (Nor, Rand, Uni). Results are consistent with the reported findings: persona-based simulators achieve the strongest ranking alignment (E-c is best on TRM, RC), while the tradeoff with distributional metrics observed on other benchmarks persists.

\section{Non-Parametric Baseline Results}
\label{sec:appendix_baselines}

For space, the main-text results (Table~\ref{tab:results}) report only the normal (Nor) non-parametric baseline. Table~\ref{tab:appendix_baselines} reports the complete set of non-parametric baselines---normal
(Nor), random (Rand, Dirichlet-Multinomial), and uniform (Uni)---for GSS, W50, and W92; the corresponding W34 baselines appear alongside its persona-source results in Table~\ref{tab:appendix_w34}.
Persona-source results for these benchmarks are given in Table~\ref{tab:results}. Here we isolate the baselines to characterize the non-parametric reference points against which persona-based simulation is
compared. The normal baseline consistently dominates random and uniform on distributional alignment (TVD, JSD, WD), motivating its use as the single representative baseline in the main text.

\begin{table}[t]
  \small
  \centering
  \setlength{\tabcolsep}{3pt}
  \caption{Simulation alignment on OpinionQA W34 across persona sources (Soc, RT, E-c) and non-parametric baselines (Nor, Rand, Uni). \underline{Underline} marks the best value; (*) indicates a statistically
significant difference from the best (paired $t$-test, $p<0.05$).}
  \begin{tabular}{lcccccc}
  \toprule
  Metric & Soc & RT & E-c & Nor & Rand & Uni \\
  \midrule
  \multicolumn{7}{l}{\textit{RADIUS (proposed alignment suite)}} \\
  TRM $\uparrow$ & \rnd{0.742} & \rnd{0.712} & \underline{\rnd{0.773}} & \rnd{0.364}* & \rnd{0.333}* & \rnd{0.318}* \\
  RC $\uparrow$  & \rnd{0.818}* & \rnd{0.810}* & \underline{\rnd{0.885}} & \rnd{0.644}* & \rnd{0.472}* & \rnd{0.334}* \\
  TVD $\downarrow$ & \rnd{0.370}* & \rnd{0.375} & \underline{\rnd{0.332}} & \rnd{0.357} & \rnd{0.468}* & \rnd{0.366} \\
  DH $\uparrow$  & \rnd{0.000} & \rnd{0.000} & \rnd{0.000} & \rnd{0.030} & \rnd{0.030} & \underline{\rnd{0.045}} \\
  \midrule
  \multicolumn{7}{l}{\textit{Prior work (for comparison)}} \\
  CV $\downarrow$ & \rnd{0.388}* & \rnd{0.396}* & \underline{\rnd{0.346}} & \rnd{0.420}* & \rnd{0.518}* & \rnd{0.451}* \\
  JSD $\downarrow$ & \rnd{0.141} & \rnd{0.149} & \rnd{0.113} & \underline{\rnd{0.110}} & \rnd{0.192}* & \rnd{0.117} \\
  WD $\downarrow$ & \rnd{0.478} & \rnd{0.523}* & \underline{\rnd{0.422}} & \rnd{0.664}* & \rnd{0.799}* & \rnd{0.693}* \\
  \midrule
  SBS $\uparrow$ & -37.9 & -38.8 & -9.0 & 2.8 & -50.1 & 0.0 \\
  \bottomrule
  \end{tabular}
  \label{tab:appendix_w34}
\end{table}

\begin{table}[t]
  \small
  \centering
  \setlength{\tabcolsep}{4pt}
  \caption{Non-parametric baseline alignment (Nor, Rand, Uni) for GSS, W50, and W92. Persona-source results (Soc, RT, E-c), and statistical comparisons against them, are reported in Table~\ref{tab:results};
W34 baselines appear in Table~\ref{tab:appendix_w34}.}
  \begin{tabular}{llcccccccc}
  \toprule
  Survey & Baseline & TRM $\uparrow$ & RC $\uparrow$ & TVD $\downarrow$ & DH $\uparrow$ & CV $\downarrow$ & JSD $\downarrow$ & WD $\downarrow$ & SBS $\uparrow$ \\
  \midrule
  \multirow{3}{*}{GSS}
    & Nor & 0.450 & 0.613 & 0.252 & 0.033 & 0.270 & 0.054 & 0.379 & -0.1 \\
    & Rand & 0.383 & 0.479 & 0.386 & 0.000 & 0.413 & 0.121 & 0.600 & -105.8 \\
    & Uni  & 0.100 & 0.172 & 0.254 & 0.017 & 0.279 & 0.051 & 0.394 & 0.0 \\
  \midrule
  \multirow{3}{*}{W50}
    & Nor & 0.391 & 0.710 & 0.380 & 0.000 & 0.424 & 0.126 & 0.688 & 7.4 \\
    & Rand & 0.312 & 0.500 & 0.492 & 0.000 & 0.527 & 0.206 & 0.861 & -25.1 \\
    & Uni  & 0.078 & 0.173 & 0.408 & 0.000 & 0.485 & 0.144 & 0.766 & 0.0 \\
  \midrule
  \multirow{3}{*}{W92}
    & Nor & 0.364 & 0.675 & 0.306 & 0.000 & 0.348 & 0.094 & 0.568 & 7.8 \\
    & Rand & 0.286 & 0.464 & 0.454 & 0.000 & 0.471 & 0.185 & 0.768 & -43.1 \\
    & Uni  & 0.299 & 0.464 & 0.331 & 0.000 & 0.424 & 0.108 & 0.612 & 0.1 \\
  \bottomrule
  \end{tabular}
  \label{tab:appendix_baselines}
  \vspace{-1mm}
\end{table}

\section{Reproducibility of \citet{Cao:2025:NAACL-HLT}}
\label{sec:appendix_reproducibility}

Section~\ref{subsec:external_complementarity} uses the fine-tuned survey simulator of \citet{Cao:2025:NAACL-HLT} as an independent, large-scale testbed for metric complementarity. Here we detail that reproduction and verify its fidelity against their published results. They fine-tune LLMs to align first-token option probabilities with World Values Survey (WVS) response distributions via a KL-divergence objective, evaluating with a Jensen--Shannon-based similarity score ($S_{JS}$) and Earth Mover's Distance (EMD). We use these two metrics only for the fidelity check below (Section~\ref{app:appendix_reproducibility:metrics}) and evaluate complementarity in the main text with RADIUS.

\subsection{Setup}

We fine-tune \texttt{DeepSeek-R1-Distill-Qwen-7B}, a state-of-the-art open model with strong performance across diverse benchmarks, using their released code, data splits (train: $C_1{\times}Q_1$, 6{,}841 samples; validation: $C_1{\times}Q_2$, 1{,}586 samples; test: 7{,}329 samples across seven country-question combinations), and hyperparameters (LoRA rank 8, $\alpha{=}32$, dropout 0.05, target modules \{q\_proj, v\_proj\}; AdamW, lr $1\mathrm{e}{-4}$). 
We train for up to 5 epochs.\footnote{All fine-tuning and inference ran on a single AWS EC2 \texttt{g5.xlarge} instance (1$\times$ A10G, 24GB). Fine-tuning took $\sim$7 hours, each inference pass $\sim$20 minutes.} Following their unmodified training script's default behavior, which checkpoints only on validation-loss improvement, without per-epoch saves, the retained checkpoint corresponds to the epoch with lowest validation loss (epoch 2 of 5 in our run), not necessarily the final epoch. We replicate their prompt construction exactly, including their custom chat-style template (a non-standard reuse of the model's \texttt{<|begin\_of\_sentence|>} token as a role separator, rather than DeepSeek-R1's native template) and their exact option-token extraction logic, as the fine-tuned checkpoint was trained on this specific format.

Two deviations from their setup remain: (1) due to GPU memory constraints (single A10G, 24GB, vs.\ their A100, 80GB) we use a training batch size of 4 with 4-bit quantization instead of their batch size of 16 at full precision; (2) their main results table averages over five of the seven test country-question conditions (6{,}673 samples), omitting two ($C_2{\times}Q_2$, $C_3{\times}Q_2$). We report both the full test set and this matching 6{,}673-sample subset.
\citet{Cao:2025:NAACL-HLT} further report that their fine-tuning yields similar performance regardless of the base model, so this choice is not critical to the comparison.

\subsection{Evaluation Metrics}
\label{app:appendix_reproducibility:metrics}

We follow the metric definitions and implementation from \citet{Cao:2025:NAACL-HLT}'s released repository, alongside RADIUS. \citet{Cao:2025:NAACL-HLT} define their similarity metric as $1 - d_{JS}(P,Q)$, where $d_{JS}$ is the Jensen--Shannon distance between the human and predicted probability vectors $P, Q$ (their code passes unnormalized percentages for $P$, which \texttt{scipy.spatial.distance.jensenshannon} normalizes internally):
\begin{equation}
    S_{JS} = 1 - d_{JS}(P,Q), \quad d_{JS}(P,Q) = \sqrt{\tfrac{1}{2}D_{\mathrm{KL}}(P\|M) + \tfrac{1}{2}D_{\mathrm{KL}}(Q\|M)}, \ M = \tfrac{1}{2}(P+Q).
\label{eq:sjs}
\end{equation}
Since this differs from the standard Jensen--Shannon divergence used elsewhere in this paper, we denote their metric $S_{JS}$ to avoid confusion.\footnote{Our JSD is the divergence itself, without the square root: $\text{JSD}(P,Q) = \tfrac{1}{2}D_{\mathrm{KL}}(P\|M) + \tfrac{1}{2}D_{\mathrm{KL}}(Q\|M) = d_{JS}(P,Q)^2$. Thus $S_{JS} \neq 1 - \text{JSD}$.} Their EMD is computed via \texttt{scipy.stats.wasserstein\_distance} applied directly to $P$ and $Q$ as unweighted sample values, rather than as probability masses over a shared ordinal support. We replicate this exactly to enable direct comparison with their published results. These two metrics are used only to verify reproduction fidelity (Table~\ref{tab:appendix_repro_comparison}). The complementarity analysis (Section~\ref{subsec:external_complementarity}) uses RADIUS's own metrics.

\begin{table}[t]
  \small
  \centering
  \caption{Our reproduction vs. \citet{Cao:2025:NAACL-HLT}'s published DeepSeek-R1-Distill-Qwen-7B results (matching test subset, $n=6{,}673$). $S_{JS}$ is their similarity metric (Eq.~\eqref{eq:sjs}), distinct from our JSD.}
  \begin{tabular}{lcccc}
        \toprule
        & \multicolumn{2}{c}{$S_{JS}$ $\uparrow$} & \multicolumn{2}{c}{EMD $\downarrow$} \\
        \cmidrule(lr){2-3} \cmidrule(lr){4-5} 
        & Paper & Ours & Paper & Ours \\
        \midrule
        Zero-shot (ZS) & 0.648 & 0.593 & 0.109 & 0.122 \\
        Fine-tuned (FT) & 0.801 & 0.797 & 0.081 & 0.080 \\
        \bottomrule
    \end{tabular}
    \label{tab:appendix_repro_comparison}
    \vspace{-1mm}
\end{table}

\begin{table}[t]
    \small
    \centering
    \caption{RADIUS metrics on full test set ($n=7{,}329$) and matching test subset ($n=6{,}673$).}
    \begin{tabular}{lcccc}
        \toprule
        & \multicolumn{2}{c}{Full test set} & \multicolumn{2}{c}{Matching test subset} \\
        \cmidrule(lr){2-3} \cmidrule(lr){4-5} 
        Metric & Zero Shot & Fine-Tuned & Zero Shot & Fine-Tuned \\
        \midrule
        TRM $\uparrow$ & 0.374 & 0.540 & 0.379 & 0.566 \\
        RC $\uparrow$  & 0.482 & 0.698 & 0.467 & 0.707 \\
        TVD $\downarrow$ & 0.498 & 0.256 & 0.498 & 0.239 \\
        DH $\uparrow$  & 0.006 & 0.029 & 0.004 & 0.031 \\
        \bottomrule
    \end{tabular}
    \label{tab:appendix_repro_radius}
    \vspace{-1mm}
\end{table}

\subsection{Results}
  
Table~\ref{tab:appendix_repro_comparison} compares our reproduction (matching test subset, $n{=}6{,}673$) to their published values. Fine-tuned scores closely match ($S_{JS}$: 0.797 vs.\ 0.801; EMD: 0.080 vs.\ 0.081), with a small residual gap in the zero-shot condition. The close match on the fine-tuned model, despite using their default best-validation checkpoint and our hardware deviations, confirms the fidelity of our reproduction.
  
Table~\ref{tab:appendix_repro_radius} reports RADIUS on the same predictions. On average, fine-tuning improves every metric: both their distributional scores and all four RADIUS metrics (TRM: 0.37$\rightarrow$0.54, RC: 0.48$\rightarrow$0.70, TVD: 0.50$\rightarrow$0.26, DH: 0.006$\rightarrow$0.029 on the full set). Taken alone, these aggregates agree with \citet{Cao:2025:NAACL-HLT}'s own metrics. 
The per-question analysis in the main text (Figure~\ref{fig:cao_scatter}) shows that this aggregate agreement nonetheless masks substantial complementarity: distribution and ranking improvements are correlated but far from equivalent, and among questions whose TVD improves, 45.9\% still misidentify the human top choice (TRM$=$0), confirming that RADIUS surfaces decision-relevant errors invisible to distributional metrics.

\end{document}